# An application of a deep learning algorithm for automatic detection of unexpected accidents under bad CCTV monitoring conditions in tunnels

Kyu Beom Lee

Department of Extreme Environment Research Center

Korea Institute of Civil Engineering and Building Technology (KICT),

University of Science & Technology (UST)

Gyeonggi-Do , The Republic of Korea

kyubeomlee@kict.re.kr

Hyu Soung Shin

Department of Extreme Environment Research Center

Korea Institute of Civil Engineering and Building Technology (KICT)

Gyeonggi-Do, The Republic of Korea

hyushin@kict.re.kr

*Abstract—* In this paper, Object Detection and Tracking System (ODTS) in combination with a well-known deep learning network, Faster Regional Convolution Neural Network (Faster R-CNN), for Object Detection and Conventional Object Tracking algorithm will be introduced and applied for automatic detection and monitoring of unexpected events on CCTVs in tunnels, which are likely to (1) Wrong-Way Driving (WWD), (2) Stop, (3) Person out of vehicle in tunnel (4) Fire. ODTS accepts a video frame in time as an input to obtain Bounding Box (BBox) results by Object Detection and compares the BBoxs of the current and previous video frames to assign a unique ID number to each moving and detected object. This system makes it possible to track a moving object in time, which is not usual to be achieved in conventional object detection frameworks. A deep learning model in ODTS was trained with a dataset of event images in tunnels to Average Precision (AP) values of 0.8479, 0.7161 and 0.9085 for target objects: Car, Person, and Fire, respectively. Then, based on trained deep learning model, the ODTS based Tunnel CCTV Accident Detection System was tested using four accident videos which including each accident. As a result, the system can detect all accidents within 10 seconds. The more important point is that the detection capacity of ODTS could be enhanced automatically without any changes in the program codes as the training dataset becomes rich.

I. INTRODUCTION

Object detection technology has been successfully applied to find the size and position of target objects appearing on images or videos. Several applications have appeared mainly in self-driving of vehicles, CCTV monitoring and security system, cancer detection, etc. Object tracking is another area in image processing to be achieved by unique identification and tracking the positions of identified objects over time. However, to track objects, it is necessary to define object class and position first in a firstly given static image by object detection. Therefore, it can be said that the results of object tracking should be deeply dependent on the performance of the object detection involved. This object tracking technology has been successfully utilized for tracing of targeted pedestrian and the moving vehicle, accident monitoring in traffic camera, criminal and security monitoring in the certain local area of concern, etc. In the traffic control field, a case study on analysis and control of traffic conditions by automatic object detection has carried out in this paper. The summaries are given as follows. According to [1], an on-road vehicle detection system for the self-driving car was developed. This system detects vehicle object and classifies the type of vehicle by Convolutional Neural Network (CNN). The vehicle object tracking algorithm tracks the vehicle object by changing the tracking center point according to the position of the recognized vehicle object on the image. Then, the monitor shows a localized image like a bird's viewpoint with the visualized vehicle objects, and the system calculates the distance between the driving car and the visualized vehicle objects. This process of the system enables to objectively view the position of the vehicle object so that it can help assistance of the self-driving system. As a result, it can localize the vehicle object in vertical 1.5m, horizontal 0.4m tolerance at the camera.

In [2], another deep learning-based detection system in combination with CNN and Support Vector Machine (SVM) was developed to monitor moving vehicles on urban roads or highways by satellite. This system extracts the feature from the satellite image through CNN using the satellite image as an input value and performs the binary classification with SVM to detect the vehicle BBox. Besides, Arinaldi, Pradana, and Gurusinga [3] developed a system to estimate the speed of the vehicle, classify vehicle type, and analyze traffic volume. This system utilizes BBox obtained by object detection based on videos or images. The algorithm applied to the system was compared with the Gaussian Mixture Model + SVM and faster RCNN. Then it appears that faster R-CNN was able to detect the position and type of vehicle more accurately. In other words, it could be said that the deep learning-based object detection approach is superior to the algorithm based object detection system. As a conclusion, all of the development cases in this paper deal with object detection based monitoring system to obtain traffic information, showing outstanding performance with deep learning. However, they all were hard to assign unique IDs to the detected objects and track them by keeping the same ID over time.

Therefore, in this paper, an attempt is made for generate an object detection & tracking system (ODTS), that can obtain moving information of target objects by combining object tracking algorithm with the deep learning-based object detection process. The full ODTS procedures (Figure 1) will be described in details in the following section. Also, the tunnel accident detection system in the framework of ODTS will be taken into consideration [4,7]. This system is used for detecting accident or unexpected events taking place on moving object and target local region on CCTV.

## II. DEEP LEARNING-BASED OBJECT DETECTION AND TRACKING SYSTEM

### A. Concept

Figure 1 shows the process of object detection and tracking by the ODTS over time[7]. It is assumed that ODTS has been trained enough to perform object detection properly on a given image frame. ODTS receives selected frames of video at specified time interval c and gains sets of coordinates, n BBoxs are detected. $BBox_T$ of objects on the given image frame at the time T, from the trained object detection system. The corresponding type or class $Class_T$ of each detected object $BBox_T$ is simultaneously classified by the object detection module.

Then, based on the detected object information, a dependent object tracking module is initiated to assign the unique ID number to each of the detected objects, $ID_T$ and predict the next position of each of the objects, $BBox'_T$. The number of tracking BBox u is different from n. But If past tracked BBox is 0, the number of tracking BBox equals to the number of the detected objects. For example, in time T+c, if u is 0, u' equals to n'. In other words, when the past tracking BBox did not exist, the current tracking BBox takes from the detected objects per each class. This object tracking module was composed by introducing an object tracking algorithm called SORT algorithm[5], which uses a concept of Intersection Over Union (IOU) to trace the same object with the same ID number and also uses Kalman filter and Hungarian algorithm for prediction of the next position of the detected objects.

At the next time step T+c, the same processes are followed on the newly given image to obtain $BBox_{T+c}$ and $Class_{T+c}$ by the same object detection module as used at the time T. Then, IOU of all the possible pairs between the predicted positions, $BBox'_{T+c}$ at time T and detected objects positions, $BBox_{T+c}$ at the time T+c are calculated. The nearest objects, namely the pair with the highest IOU value, will be assumed the same object with the same ID. And any object in $BBox'_T$ which has no object pair with higher than IOU value of 0.3 will be considered as disappeared from the region of interest (ROI).

Similarly, any object in $BBox'_{T+c}$ which has no object pair with higher than IOU value of 0.3 will be considered as newly appeared into the RoI at T+c. The freshly emerged object will be assigned by a new ID number not overlapped with the previous ID number.

This system utilizes a faster RCNN learning algorithm[5] for object detection and a SORT[6] for ID assignment and object tracking. It is known that SORT[6] enables multi-object tracking by 100-300fps degree speed. These system processes object tracking using SORT[6] algorithm based on IoU value, so the object tracking ability was affected by video frame interval c[7]. Video frame interval can reduce the computation amount over time by adjusting the detection interval of the object detection network. To check this, object tracking ability over the frame interval experimented, and then it was possible to track the objects until six frame interval[7]. Increasing frame interval significantly reduces object tracking ability, so that the video frame interval should be optimized for the number of camera devices simultaneously connected to a deep-learning server.

### B. Tunnel accident detection system

Driving in a road tunnel is dangerous because it is inadequate space to evacuate compared to a general highway, and should inform the drivers if an emergency occurs in the tunnel[7]. Therefore, in South Korea, Person, Fire, Stop, and

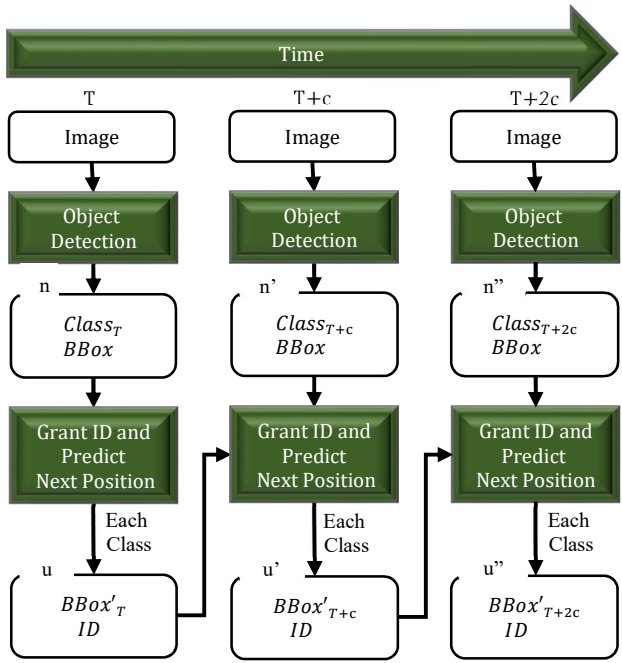

Figure 1. Object detection and tracking process of Object Detection-Tracking system over time. When class and BBox were obtained by object detection, object tracking algorithm grants ID and predicts next position using current and past BBox.

Wrong Way Driving(WWD) are the target objects and events to be monitored in the national regulation[4]. Driving in a road tunnel is dangerous because it is inadequate space to evacuate compared to a general highway. Therefore, drivers should be informed as soon as possible when an emergency occurs in the tunnel[7]. In South Korea, The objects or events: Person, Fire, Stop, and WWD should be detected and monitored according to the Korean national regulation [4].

Meanwhile, The monitoring of the target objects and unexpected events are undertaken through CCTVs in tunnels. And an automatic object detection system for the targets would be adopted for the purpose with an excellent performance outside of tunnels. However, the system doesn't work at all in a tunnel. It is because: (1) In the tunnel video has low Illuminance, so the video was greatly influenced by the tail light of the driving vehicle or the warning light of the car in operation. (2) The tone of the tunnel video was a dark color. In other words, it has a different color compared to the road of the tunnel outside. Since the above two reasons, the video monitoring system developed on the roads outside of the tunnels was likely to fail to operate appropriately in the tunnel. Therefore, an automatic accident detection system specialized for road tunnel is required.

To overcome the problems above, deep learning-based Tunnel CCTV Accident Detection System was developed in [7]. A deep learning model of Faster R-CNN was used for training. And this model was based on a model that learns image datasets that include some accidents in tunnels. Then, ODTS uses an object tracking function only with Car object, and the tracking information of the target Car object is used for determining Stop and WWD events by using Car Accident Detection Algorithm(CADA) periodically.

The procedure of CADA detects an accident state, as shown in Figure 2. First, set Region of Interest(RoI) to on CCTV screen in the tunnel and proceed crop and warp to the extracted image from the ROI of the original image of a CCTV screen. This procedure is similar to [1], but the purpose of this system is to achieve a coherent standard for identifying Stop and WWD events. The image extract makes the image more comfortable to be trained by removing unnecessary image area out of ROI and by making a similar size of objects of the near and far side. These points are different from [1]. Next, detect Car, Fire, Person objects by trained Faster R-CNN[5].

Next, To prevent false detection of the Fire object, Extra 'No Fire' object was defined by directly defining the object class to reduce the wrong answer for Fire object. The No Fire object is assigned for misleading objects like tunnel light, tail light of the car, etc. The feature of those data is reflected in training that designates the pre-defined object class different from the exception of background in Faster R-CNN training. With this method, the reduction of Fire misdetection on untrained data could be possible.

Then, the Car object is numbered by object tracking algorithm, and the previous cycle BBox of the same ID is compared with the current cycle BBox at a predetermined

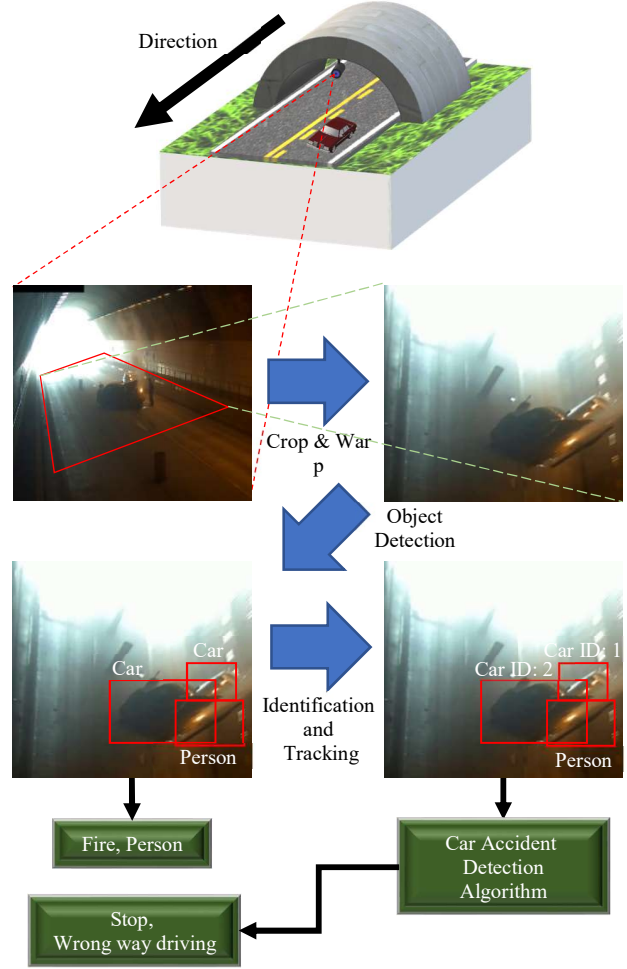

Figure 2. Accident detection process using tunnel CCTV. Object detection can find Fire, Person accident and object tracking finds Stop and Wrong way driving accident

time interval with CADA. At this time, WWD was decided by Intersection over Line(IoL), and Stop was decided by IoU. IoL is a concept, the ratio of overlapping lines similar to IoU. Use only the vertical value of BBox to determine inversion, as shown in the following equation.

$$\text{IoL} = \frac{Overlapped\ Length\ of\ Vertical\ element\ of\ BBox}{Union\ Length\ of\ Vertical\ element\ of\ BBox} \quad (1)$$

The criterion of WWD is based on the warped tunnel CCTV image to detect the vertical velocity of the BBox in the image. If IoL is less than 0.75 and the direction of vertical driving is the opposite, it was judged to be inverse.

IOU was used as a criterion for determining the Stop event because the position and size of the BBox should be considered regardless of the direction. If it is 0.9 or more, it is judged as a stop.

Through the process as mentioned above, the tunnel CCTV video accident detection system is applied for an

9

experiment with real object images and event videos obtained from tunnel monitoring sites.

## III. Experiments

Experiments with the developed system in this study are divided into two parts: the learning performance measurement of deep learning and the accident detection performance of the entire system. The SORT used in ODTS is greatly affected by object recognition performance. Therefore, to complete this system, high performance of object detection through proper learning of deep learning object detection network was required.

Then, based on the trained deep learning model, the entire system was tested to see if it can detect the targeted four accident events. In this case, since both the object detection performance of the deep learning model and the discriminative ability of the CADA were both required, the system was tested for each image to determine whether it is possible to detect each situation.

### A. Deep learning training

Training of the deep learning network was undertaken by not a video but a series of still images. In this paper, a single cycle of the training process for a whole dataset is defined by one epoch. The dataset to be learned involves the images in accident events. Faster R-CNN[5] was used for training.

TABLE I. THE STATUS OF USED IMAGE DATASET

| Number of Videos | Number of images | Number of objects | | |
|---|---|---|---|---|
| | | Car | Fire | Person |
| 45 | 70914 | 427554 | 857 | 44862 |

Table.1 shows the status of the dataset used for training. This dataset is composed of 70,914 video images by dividing 45 videos into frames. Unlike general deep learning process, learning data and inference data were not separated in the training process of deep learning. This is because, unlike the publicly available datasets, the dataset used in this paper is that the images are continuous in each video. In other words, the images present in each video file have the same image background and differ depending on the presence of the objects. If the training data and the inference data were divided for each image, the inference performance of the object detection network would show similar performance. On the other hand, the stability of the object detection on the whole video may be deteriorated, which adversely affects the detection performance of the accident, so it is difficult to test the detection procedure of the entire tunnel CCTV image accident detection system. Therefore, training was performed by collecting all available data, and evaluation of the deep learning object detection performance uses the learned data.

The number of Fire objects is a little because the occurrence of fire events in the tunnel is infrequent to happen. Therefore, there is a high possibility of false detection and missed detection for fire, and what is vital in the tunnel control center is that false detection should be fewer than missed detection.

If the system installed in the field is frequently informed that false detection has occurred even in the absence of false detection, the reliability of the system is substantially degraded. On the other hand, in the case where the data was not detected, it was possible to improve the detection performance automatically with the enriched dataset in the time-lapse which is periodically included in the training dataset. Therefore, the experiment focused on reducing false detection, so that the number of No Fire objects were much higher than these of fires.

10 epochs proceeded faster R-CNN training. The deep learning framework was Tensorflow 1.3.0 on Linux[7]. The hardware of Faster R-CNN training uses Nvidia GTX 1070. The training time took 60 hours, and the inference performance of each object class evaluates by Average precision(AP).

TABLE II. INFERENCE RESULT OF DATASET

| Number of images | Average Precision (AP) | | |
|---|---|---|---|
| | Car | Person | Fire |
| 70914 | 0.8479 | 0.7161 | 0.9085 |

AP values for the three target objects to be detected were shown in Table.2. in the training dataset, the number of Car objects is the largest object and very high AP value was obtained for the Car object in comparison with other classes. That is, the object detection performance of deep running of the Car in the video was expected to be highly reliable. On the other hand, in Table.2, AP for Person object results in relatively low value because Person object exists long, tiny shape in small size.

The AP of Fire object was high as 0.9085, but false detection for the object might be highly possible as the number of the objects available for training was very small, 857. Nonetheless, training about deep learning, including No Fire objects, could reduce the false detection of Fire object. However, to detect the Fire in the tunnel control center, it was necessary to collect and involve more images of a Fire event in training.

### B. Accident detection test using entire Tunnel CCTV Accident Detection System

Based on the trained deep learning model, the performance in accident detection of the deep learning-based Tunnel CCTV Accident Detection System needs to be evaluated. For this, 4 videos were selected to check up for 4 events as in

Table 3. A visualization program was generated to show detection results on the video.

The video frame interval was set to 6 frames per second at 30 fps, and it was evaluated that it was detected within 10 seconds after visual observation[7]. The length of the video, the time of occurrence, and the detected time are summarized in Table 3.

TABLE III.  DETCTIED TIME OF THE EACH ACCIDENT BY ACCIDENT DETECT SYSTEM

| Accident video information | Item on video time | | |
|---|---|---|---|
| | *Video length* | *Occurrence time* | *Detected time* |
| Stop | 126s | 5s | 7s |
| Wrong Way Driving | 29s | 4s | 12s |
| Fire | 64s | 29s | 29s |
| person | 72s | 50s | 50s |

In Table 3, there is a time difference between occurrence and the detection of Stop and WWD events. Since this is a characteristic of CADA, in our experiment, it is detected every 2.4-sec cycle. Nevertheless, the system could detect the difference of 2 seconds for Stop and 8 seconds for WWD. On the other hand, images, including Person and Fire, showed rapid detection immediately after the accident. However, there is a limitation that the images used in Table 3 are images used for training so that it will be different from the case where it was installed directly in the field. Therefore, the application of the testbed and additional videos for testing were required.

IV. CONCLUSION

This paper proposes a new process of ODTS by combining deep learning-based object detection network and object tracking algorithm, and it shows dynamic information of an object for a specific object class can be obtained and utilized.

On the other hand, the object detection performance is important because SORT used in ODTS object tracking uses only information of BBox without using an image. Therefore, continuous object detection performance may be less needed unless the object tracking algorithm is relatively dependent on object recognition performance.

And Tunnel CCTV Accident Detection System based on ODTS was developed. The experiments on training and evaluation of deep learning object detection network and detection of an accident of the whole system were conducted. This system adds CADA that discriminates every cycle based on dynamic information of the car objects. As a result of experimenting with the image containing each accident, it was possible to detect the accidents within 10 seconds.

On the other hand, training of deep learning secured the object detection performance of a reliable Car object, and Person showed relatively low object detection performance. However, in the case of Fire, there is a high probability of false detection in the untrained videos due to the insufficient number of Fire objects. Nonetheless, it is possible to reduce the occurrence of false detections by simultaneously training objects that are No Fire. The fire object detection performance of the deep learning object detection network should be improved by securing the Fire image later.

Although the ODTS can be applied as an example of a Tunnel CCTV Accident Detection System, it is also used to fields that need to monitor the dynamic movement of a specific object such as vehicle speed estimation or illegal parking monitoring will be possible. To increase the reliability of the system, it is necessary to secure various images and to secure Fire and Person objects. Besides, through the application and continuous monitoring of the tunnel management site, the reliability of the system could be improved.

REFERENCES

[1] E. S. Lee, W. Choi, D. Kum, "Bird's eye view localization of surrounding vehicles :Longitudinal and lateral distance estimation with partial appearance," Robotics and Autonomous Systems, 2019, vol. 112, pp. 178-189.

[2] L. Cao, Q. Jiang, M. Cheng, C. Wang, "Robust vehicle detection by combining deep features with exemplar classification," Neurocomputing, 2016, vol. 215, pp. 225-231.

[3] A. Arinaldi, J. A. Pradana, A. A. Gurusinga, "Detection and classification of vehicles for traffic video analytics," Procedia computer science, 2018, vol. 144, pp. 259-268.

[4] K. B. Lee, H. S. Shin, D. G. Kim, "Development of a deep-learning based automatic tunnel incident detection system on cctvs," in Proc. Fourth International Symposium on Computational Geomechanics, 2018, pp. 140-141.

[5] S. Ren, K. He, R. Girshick, J. Sun, "Faster R-CNN: Towards Real-Time Object Detection with Region Proposal Networks," in Proc. Neural Information Processing Systems, 2015, pp. 91-99.

[6] A. Bewley, Z. Zongyuan, L. Ott, F. Ramos, B. Upcroft, "Simple Online and Realtime Tracking," in Proc. IEEE International Conference on Image Processing, 2016, pp. 3464-3468.

[7] K. B. Lee, H. S. Shin, D. G. Kim, "Development of a deep-learning based automatic tracking of moving vehicles and incident detection processes on tunnels," Korean Tunnelling and Underground Space Association, 2018, vol. 20, no.6, pp. 1161-1175.